# Context Aware Dynamic Traffic Signal Optimization

Kandarp Khandwala
VESIT, University of Mumbai
Mumbai, India

kandarpck@gmail.com

Rudra Sharma
VESIT, University of Mumbai
Mumbai, India

rudrsharma@gmail.com

Snehal Rao
VESIT, University of Mumbai
Mumbai, India

snehalrao09@gmail.com

## ABSTRACT

Conventional urban traffic control systems have been based on historical traffic data. Later advancements made use of detectors, which enabled the gathering of real-time traffic data, in order to re-organize and calibrate traffic signalization programs. Further evolution provided the ability to forecast traffic conditions, in order to develop traffic signalization programs and strategies pre-computed and applied at the most appropriate time frame for the optimal control of the current traffic conditions. We, propose the next generation of traffic control systems based on principles of Artificial Intelligence and Context Awareness. Most of the existing algorithms use average waiting time or length of the queue to assess an algorithm's performance. However, a low average waiting time may come at the cost of delaying other vehicles indefinitely. In our algorithm, besides the vehicle queue, we use 'fairness' also as an important performance metric to assess an algorithm's performance.

## General Terms

Optimization, Intelligent Systems, Algorithms

## Keywords

Location, Context, Artificial Intelligence, Pattern, Analysis, Maps, Traffic, Optimization, Accident

## 1. INTRODUCTION

### Motivation

With every passing day, traffic snarls at major junctions are posing a great problem to roadway commuters. This can be largely attributed to lack of systematic lights, and specifically ideal light cycles is a crucial task in present day cities with potential non-profits regarding vitality utilization, traffic flow management, pedestrian safety, and environmental issues. In any case, not many productions in the current writing handle this issue by means of automatic intelligent systems, and, when they do, they focus on limited areas with elementary traffic light schedules. Hence, the pressing need for a context aware traffic management system is increasingly growing. In this work, we hence state and explain the design of an algorithm which optimizes signal timing by weighing various factors like vehicular density, waiting time at each junction, popularity of the road/street and the number of lanes on the road.

### 1.2 Problem Definition

The primary objective of our algorithm is to simplify and reduce urban traffic issues at major intersections by proposing a solution which is influenced by the present traffic conditions i.e. queued traffic, vehicular density, waiting time and need to address emergency situations like ambulance/fire brigade with highest priority. The algorithm estimates which signal should turn green and for what duration depending on the traffic situation and various factors mentioned above.

## 1.3 Relevance of the project

The algorithm so proposed can be deployed at various traffic control centres to control the timing of traffic signals at busy intersections. This could prove to be highly useful for solving traffic snarls encountered at most of the busy junctions currently fitted with traffic signals working with static signal timing. Our context aware algorithm can adapt to current traffic conditions and propose an optimal solution accordingly. This would mean reduced traffic issues, less travel time, more fuel efficiency, less pollution and happy roadway commuters.

## 2. LITERATURE SURVEY

Urban traffic control systems evolved through three generations. [1] The first generation of such systems has been based on historical traffic data. The second generation took advantage of detectors, which enabled the collection of real-time traffic data, in order to re-adjust and select traffic signalization programs. The third generation provides the ability to forecast traffic conditions, in order to have traffic signalization programs and strategies pre-computed and applied at the most appropriate time frame for the optimal control of the current traffic conditions. Nowadays, the fourth generation of traffic control systems is already under development, based among others on principles of artificial intelligence and having capabilities of on-time information provision, traffic forecasting and incident detection is being developed according to principles of large-scale integrated systems engineering. Although these systems are largely benefiting from the developments of various information technology and computer science sectors, it is obvious that their performance is always related to that of the underlying optimization and control methods. Until recently, static traffic assignment (route choice) modes were used in order to forecast future traffic flows.[2] Most of the papers surveyed use average waiting time or length of the queue to assess an algorithm's performance. However ,a low average waiting time may come at the cost of delaying other vehicles indefinitely. In our algorithm, besides the vehicle queue, we use 'fairness' also as an important performance metric to assess an algorithms performance.

## 3. DESIGN

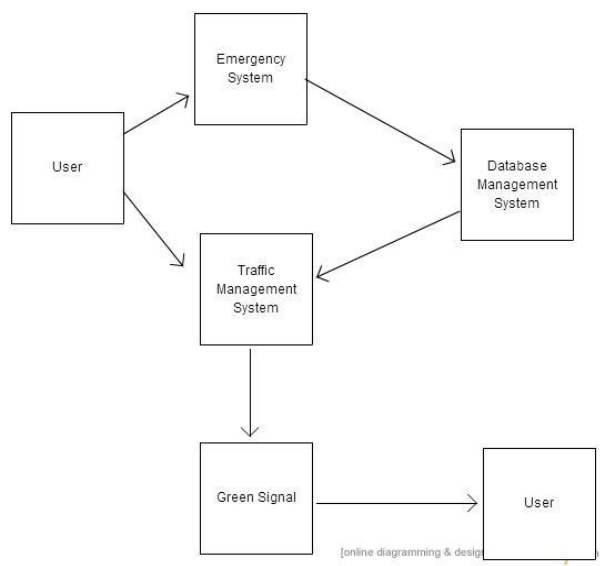

**Figure 1. System Block Diagram**

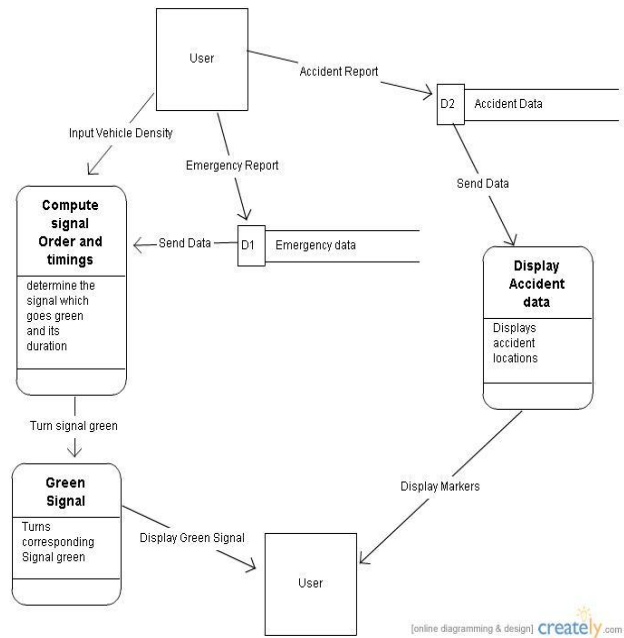

**Figure 2. Data Flow Diagram**

## 4. IMPLEMENTATION

### Data Collection

This project required a tremendous amount of data to be collected. This was accomplished by using an Android application to gather Geo-location data. Updates were sent every five seconds from the test devices when the user was traveling in a car.

The data collected included the following elements:
- Existing signal timings and operational settings
- Area Photography
- Average Vehicle Density
- Average Stop Time
- Average number of signal cycles required for the vehicle to pass
- Intersection details
- Weather Data
- Emergency Reports
- Accident Reports

### Report Generation

The data collected for each corridor was then summarized and a report was generated analysing the data in hand. The accuracy of the data collection was crucial. Some filters were also put in place to remove irrelevant data.

### Plan Formation

New timing plans were developed for each corridor using the data previously collected.
Two plans were developed for each area; peak and off peak time of the day. Consideration was also given to seasonal, special events, emergencies, rallies taking place on the road at that particular time where appropriate.

### Algorithm Design

After the cycle lengths were determined, the optimal phase times were established. This data was coded into the Instamaps Algorithm and the design was finalized.

Algorithm was designed to minimize the vehicle density, thus reducing vehicle stops and delay.

## Generalization

The penultimate step was to prepare the Instamaps algorithm for a generalized use case. The data recorded was used to develop three general cases - Low, Medium and High which denote a fuzzy representation of the current traffic conditions in this part of Mumbai, an urban area. Various tweaks were added to keep a minimum and maximum threshold, driver satisfaction and uniformity.

## Testing

Several test cases were tried and modifications were applied to the algorithm. Safeguards were added to prevent accidents and the algorithm was made crash resistant.

# 5. CASE STUDY

## Case Study - Amar Mahal Junction

One of the busiest junctions on the Eastern Express Highway - Amar Mahal Junction is located in the heart of Mumbai, India. It is a major arterial road connecting places like Ghatkopar, Vikhroli, Chembur, Santacruz, Mankhurd, Vashi and Kurla. The Mumbai Metropolitan Region Development Authority (MMRDA) manages the roads along with the Amar Mahal Junction flyover which consists of three flyovers and ten lanes. It consists of link roads, highways and station access roads. The average waiting time widely varies throughout the day. From free flowing traffic in the morning hours, the traffic comes to a grinding halt during peak office hours (9am - 11am and 5pm - 8pm).

However, all this while, the traffic signal time remains constant. It is 60 seconds for the signal connecting Ghatkopar and Chembur while just 30 seconds for the traffic coming from Kurla or Vikhroli. It leads to unpredictable traffic and huge jams. Cars have to wait for at least 4 signal cycles to pass the junction and the stop time can rise up to 5-8 minutes. The peak vehicle density is close to 200 is one direction.

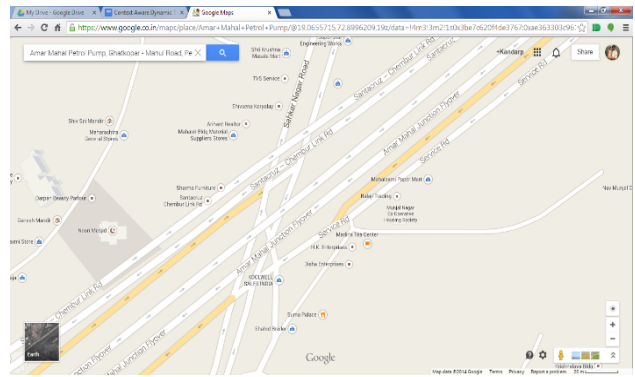

**Figure 3. Amar Mahal Junction**

Our algorithm can handle this situation efficiently. It intelligently senses the context and manages the conditions. During our testing conditions, the wait time for north - south bound vehicles was reduced by 10% during peak times by effectively routing traffic. During non-peak hours, the load was effectively balanced with the algorithm working towards attaining the optimal timing for each lane. Vehicle density per lane per day were reduced by a factor of 5% by routing to alternate paths.

# 6. SIMULATOR

We developed a simulator to test the various cases

**Figure 4. Moderate fuzzy traffic inflow and outflow at a junction**

The above Figure represents user interface to accept initial data required by the algorithm such as vehicle density on all the 4 roads at the junction and the range for the number of incoming/outgoing vehicles for each of the four roads depending on its popularity, etc.

**Figure 5. Example report generated using the above data Depending on the input provided by the user, the algorithm provides a context aware traffic report indicating which signal should turn green and for what duration**

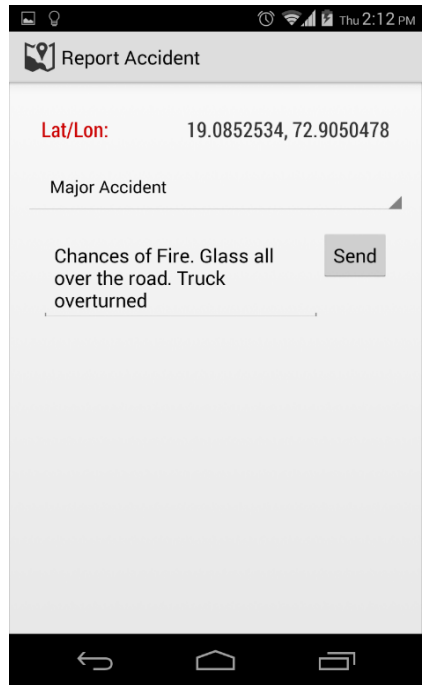

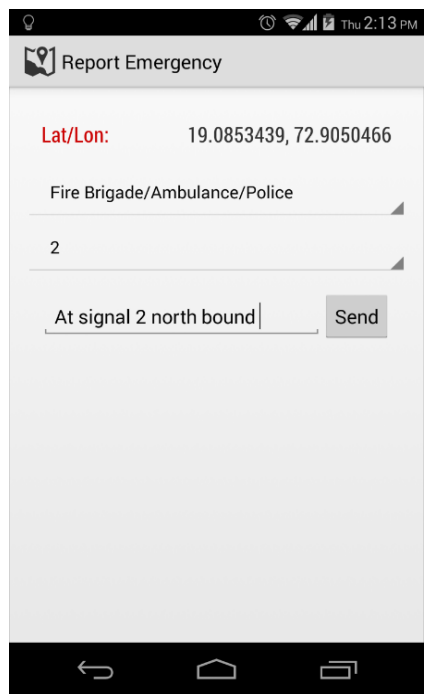

**Figure 6 & 7. Report accident and traffic from the Mobile application**

       Roadway commuters can make use of the Android application designed by us to alert the traffic control center about accidents witnessed by them so that traffic can be navigated through an alternative route if necessary. Also, people owning this Android app installed may view the accident sites on a map and accordingly take routing decisions to avoid chaos.

**Figure 8 & 9. Snapshots of the Emergency and Accident Reports**

## 7. GRAPHICAL ANALYSIS

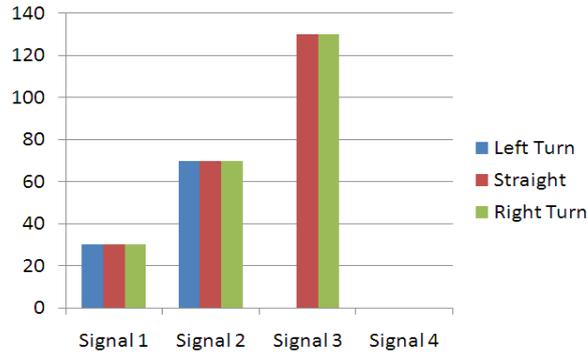

**Figure 10. Instamaps Algorithm: Representation of stoppage time at each of the signals for each direction**

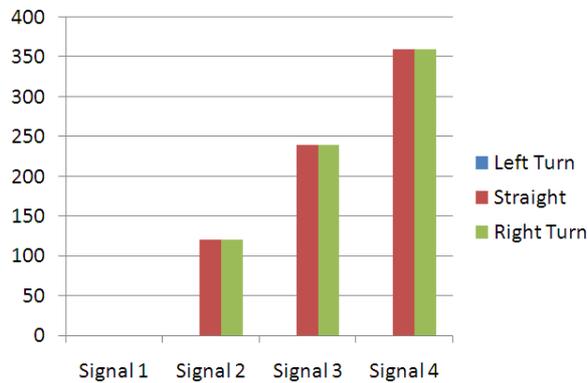

**Figure 11. Current Static Algorithm: Above graph represents the stoppage time at each of the signals for each direction (Cycle of 120 seconds has been considered in the above depiction)**

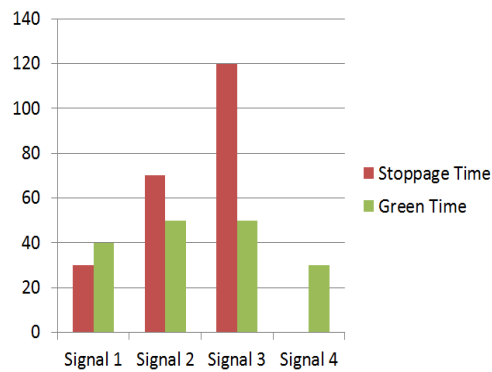

**Figure 12: Above graph depicts the stoppage time and the green time at each and every signal according to our proposed dynamic context-aware traffic optimization algorithm .The ordering of signals dynamically is 4-1-2-3.**

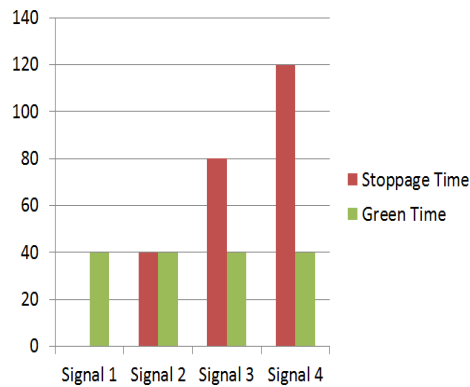

**Figure 13:Above graph depicts the stoppage time and green time at each and every signal according to the current static traffic system .The ordering of signals statically is 1-2-3-4**

## 8. OBSERVATION

Our algorithm intelligently senses the context and manages the traffic conditions. As observed from Figure 10 and Figure11, maximum waiting time at a red signal in a dynamic system is 130 seconds as compared to 360 seconds in a static system. Also, the order in which the signal turns green depends upon the road having maximum vehicle density and is not predetermined and set to signal 1 as seen in Figure 11.If the static system was to be implemented,the vehicles needing highest preference for the green signal i.e. .signal 4 would have to wait for the longest time thus rendering it inefficient.

As observed from figure 12 and 13, the stoppage time steadily goes on increasing in the static system and so does the gap between the green time and the stoppage time(Figure 13).In case of the dynamic system(Figure 12),the green time varies proportionally with the stoppage time so that the vehicles lining up at one end can be provided with sufficient time to clear up. If the static system was to be implemented, the vehicles needing highest preference for the green signal i.e. .signal 4 would have to wait for the longest time, thus rendering it inefficient.

## 9. REFERENCES

[1] http://www.sciencedirect.com/science/article/pii/S1877042811014297

[2] http://nsl.csie.nctu.edu.tw/files/ITSC2012.pdf

[3] SrinivasaSunkari, P.E., "The Benefits of Retiming Traffic Signals," ITE Journal, April 2004, pages 26-29.

[4] Traffic Signal Timing Optimization Study for Metro Nashville Signal System

[5] Salama, A.S.; Saleh, B.K.; Eassa, M.M., "Intelligent cross road traffic management system (ICRTMS)," Computer Technology and Development (ICCTD), 2010 2nd International Conference on , vol., no., pp.27,31, 2-4 Nov. 2010doi: 10.1109/ICCTD.2010.5646059